\documentclass[11pt,a4paper]{article}
\usepackage[hyperref]{ranlp2023}
\usepackage{times}
\usepackage{latexsym}

\usepackage{latexsym,graphicx,xurl}
\usepackage{algorithm}
\usepackage{algpseudocode,amsmath}

\usepackage{microtype}

\aclfinalcopy %

\title{\textsc{Vocab-Expander}: A System for Creating Domain-Specific Vocabularies Based on Word Embeddings}  %

\author{Michael Färber \\
 Karlsruhe Institute of Technology (KIT) \\
 Institute AIFB \\
 Karlsruhe, Germany \\
  {\tt michael.faerber@kit.edu} \\\And
  Nicholas Popovic \\
 Karlsruhe Institute of Technology (KIT) \\
 Institute AIFB \\
 Karlsruhe, Germany \\
  {\tt popovic@kit.edu}}
\date{}

\begin{document}
\maketitle
\begin{abstract}
In this paper, we propose \textsc{Vocab-Expander} at \url{https://vocab-expander.com}, an online tool that enables end-users (e.g., technology scouts) to create and expand a vocabulary of their domain of interest. It utilizes an ensemble of state-of-the-art word embedding techniques based on web text and ConceptNet, a common-sense knowledge base, to suggest related terms for already given terms. The system has an easy-to-use interface that allows users to quickly confirm or reject term suggestions. 
\textsc{Vocab-Expander} offers a variety of potential use cases, such as improving concept-based information retrieval in technology and innovation management, enhancing communication and collaboration within organizations or interdisciplinary projects, and creating vocabularies for specific courses in education. 
\end{abstract}

\section{Introduction}

\textbf{Motivation.} %
In many scenarios, it is necessary to create an ontology or other formal model of a domain of interest from scratch. For instance, in the field of technology and innovation management, technology scouts and other end-users without technical skills often use a list of terms for continuously retrieving and scanning texts from different media sources (e.g., news articles, social media, publications, patents) in order to become aware of novel relevant technologies and to create and populate profiles of technologies and actors within a particular domain, such as Smart Cities. However, coming up with such a vocabulary is typically highly time-consuming and costly due to the domain-specificity (i.e., non-experts have no starting point what to add), the 
complexity of correctly defining the scope
of the domain (e.g., Smart Cities can range from Smart Home to energy efficiency to security), the ambiguity of natural language (i.e., the meaning of terms may vary depending on the context in which they are used), and the emergence of new terms over time.

\begin{figure*}[tbp]
    \centering
    \includegraphics[width=\linewidth]{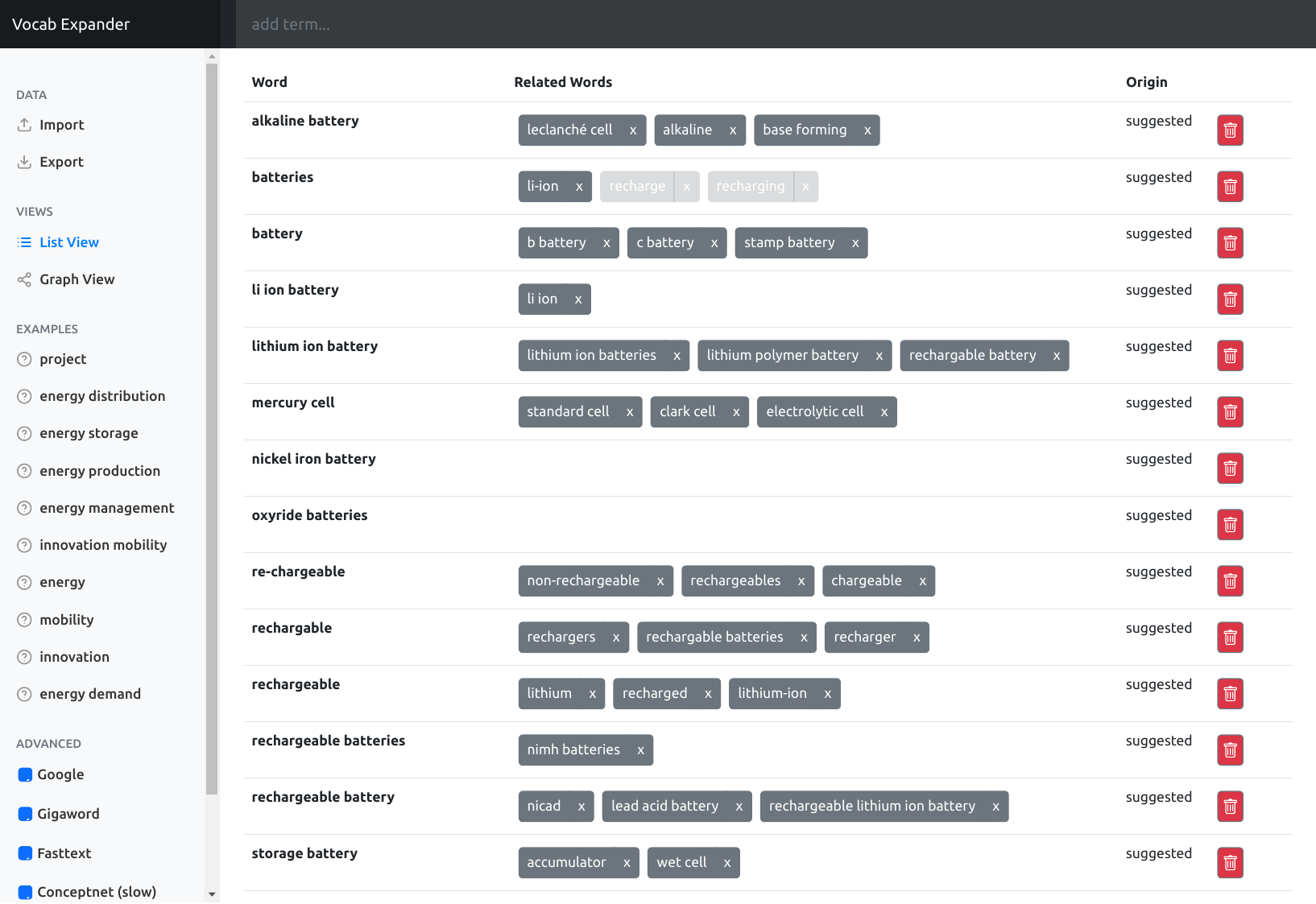} %
    \caption{Screenshot of the \textsc{Vocab-Expander}, available at \url{https://vocab-expander.com}.}
    \label{fig:screenshot}
\end{figure*}

\textbf{Current Situation.} 
So far, domain experts (e.g., technology scouts in technology and innovation management) still rely heavily on domain expert knowledge \cite{de2022technology}. Several tools for modeling a domain of interest exist, including Protegé \cite{DBLP:journals/aimatters/Musen15} and D-Terminer \cite{rigouts2022d}. 
However, these tools are often considered as ``too heavy'' for creating only a domain-specific vocabulary instead of an ontology with a specific data model and standardizations (e.g., W3C RDF, OWL). Furthermore, these tools are typically designed to support the modeling process (e.g., based on an existing text corpus), but do not suggest directly related terms for given terms.

\begin{figure*}[tbp]
    \centering
    \includegraphics[width=0.8\linewidth]{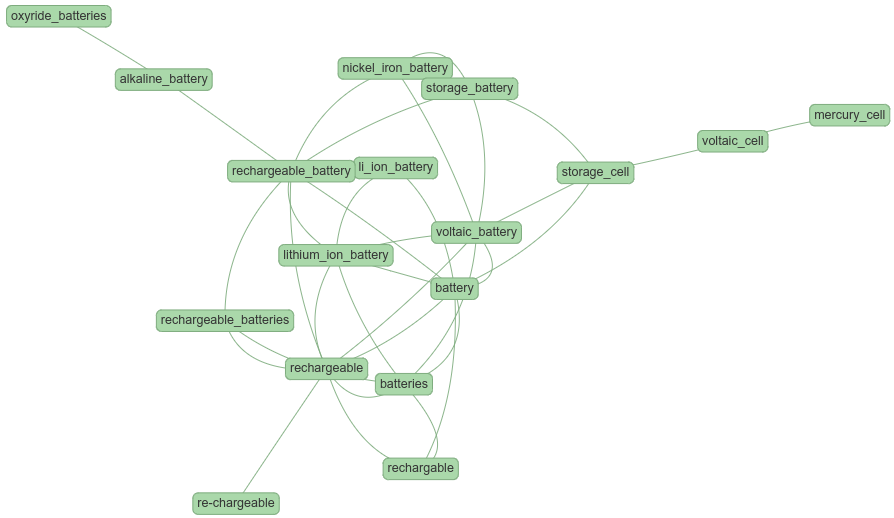} %
    \caption{Screenshot}
    \label{fig:graph}
\end{figure*}

\textbf{Contributions.} 
In this paper, we propose the system \textsc{Vocab-Expander}, available online at \url{https://vocab-expander.com}, that enables end-users without technical skills to create and expand a vocabulary of their domain of interest. 
The system utilizes an ensemble of state-of-the-art word embedding techniques to suggest related terms for already given terms. In addition to word embedding models based on web text, the system also incorporates embeddings based on ConceptNet, a common-sense knowledge base. The system is equipped with an easy-to-use interface that allows end-users to quickly confirm or reject term suggestions. The ranking of the suggested terms is based on the number of links they possess to other terms within the vocabulary. 
The created vocabulary can be listed as a table and visualized as a graph (see Figures \ref{fig:screenshot} and ~\ref{fig:graph}). We also provide an import and export functionality for the vocabularies.

\textbf{Use Cases.} Our tool offers a variety of potential use cases. For tasks such as technology and innovation management, it can be used to improve concept-based information retrieval by utilizing the created domain-specific vocabulary as search terms. Additionally, the created vocabulary can serve as a basis to enhance communication and collaboration within organizations or interdisciplinary projects by ensuring the use of consistent terminology among all involved parties. 
In the field of education, our tool allows for the creation of vocabularies for specific courses or subjects, ensuring that all relevant terms within a field or subject are covered. 
Overall, our system provides a valuable solution for creating and maintaining domain-specific vocabularies, which can be used in various fields to improve information retrieval, human communication, and natural language processing.

\textbf{Provisioning.} The source code of our system is publicly available on GitHub (\url{https://github.com/nicpopovic/VocabExpander}) under the MIT License, making it easy to reuse and adapt for a wide range of use cases.

\section{System Design}

\if{0}
The system utilizes an ensemble $E$ of state-of-the-art pre-trained word embedding models available in gensim \cite{rehurek_lrec} to suggest related terms for already given terms. Specifically, the user can choose one or several of the following models:
\begin{itemize}\itemsep0em 
 \item "word2vec-google-news-300" \cite{mikolov2013efficient}%
 \item "glove-wiki-gigaword-300" \cite{pennington2014glove}%
 \item "fasttext-wiki-news-subwords-300" \cite{mikolov2018advances}%
 \item "conceptnet-numberbatch-17-06-300" \cite{speer2017conceptnet}%
\end{itemize}
\fi

The system utilizes an ensemble $E$ of state-of-the-art pre-trained word embedding models available in \textit{gensim} \cite{rehurek_lrec} to suggest related terms for already given terms. Specifically, the user can choose one or several of the following models: (1)~\textit{word2vec-google-news-300} \cite{mikolov2013efficient}, (2)~\textit{glove-wiki-gigaword-300} \cite{pennington2014glove}, (3)~\textit{fasttext-wiki-news-subwords-300} \cite{mikolov2018advances}, (4)~\textit{conceptnet-numberbatch-17-06-300} \cite{speer2017conceptnet}.

Words $w \in W$ are categorized into 3 categories, accepted words $W_a$, rejected words $W_r$, and suggested words $W_s$. 
Initially, a user adds one or more words to $W_a$.
For each word $w_a \in W_a$ the top $k$ most similar words $w_{sim} \in W_{sim}$ according to each embedding model $e \in E$ are fetched along with the average pairwise similarity scores $P_{w_{sim, j}, w_{i}}$ across $E$.
$w_{sim} \notin W$ are added to $W_s$.
Next, we calculate a score $S_{w_s}$ for each suggested word $w_s \in W_s$ by aggregating similarity scores to accepted words and subtracting weighted similarity scores to rejected words:
\begin{equation*}
S_{w_{s,i}} = \sum_{w_{a,j} \in W_a} P_{w_{s,i}, w_{a,j}} - \lambda \sum_{w_{r,k} \in W_r} P_{w_{s,i}, w_{r,k}}
\end{equation*}
where $\lambda = 0.5$. 
Suggested words are then associated with the accepted word with which they have the highest pairwise similarity and ordered according to their score $S_{w_{s,i}}$.

The system's frontend, presented in Figure \ref{fig:screenshot}, displays the list view of accepted words and the corresponding suggested words. 
The list view showcases the three highest-ranked suggestions for a selected accepted word. 
If the score of a suggested word falls below a pre-determined threshold, a lower opacity indicates this.
Users can quickly accept a suggested word by clicking on it or reject it by clicking the "x" button next to it. 
Additionally, a graph view is available as shown in Figure \ref{fig:graph}, allowing users to visualize the similarity scores between accepted words. 
The user interface also includes import and export buttons in the top left corner, enabling the import and export of vocabulary lists.

\section{Related Work}

\textbf{Ontology Engineering and Ontology Learning.} %
Various methods have been proposed for constructing an ontology for a specific domain in a manual, semi-automated, or automated way \cite{hazman2011survey}. Automated methods typically involve extracting concepts and relations between them from domain-specific text corpora provided by the user \cite{DBLP:conf/hicss/ElnagarYT20}. 
In contrast to them, our approach does not rely on the availability of a large text corpus; instead, we enable users (domain experts as well as newcomers) to independently explore and discover related concepts from scratch. Furthermore, ontology learning \cite{buitelaar2005ontology} typically includes additional processing steps, which are out of our scope, such as clustering the concepts with identical or similar meanings and assigning unique identifiers to concepts.

\textbf{Automated Term Extraction}. 
Research on automatically extracting terms from text corpora, such as named entities, has been performed extensively. Early approaches on automatic term extraction combined linguistic hints, e.g., part-of-speech patterns, with statistical metrics for calculating the termhood and unithood \cite{kageura1996methods}, which allows to quantify to which degree the candidate term is related to the domain. Rule-based approaches have been used through many years (e.g., \citet{daille1994approche,drouin2003term}) and are still popular nowadays \cite{kosa2020optimized}. 
Machine learning-based approaches for automated term extraction 
utilize, among other things, external data sets and web search \cite{ramisch2010mwetoolkit} and word embeddings \cite{wang2016featureless,amjadian2018distributed}. Newest approaches are also based on 
language models (e.g., \cite{gao2019feature,lang2021transforming}), but require, as many other approaches, more context than a few keywords as input as for our system. 

\textbf{Demo Systems for Automated Term Extraction.} \citet{rigouts2022d} proposed D-Terminer, a running system for monolignual and bilingual automatic term extraction. In contrast to us, they focus on multiple languages, and use a text corpus as input for the system. Additionally, TermoStat \cite{drouin2003term} and TerMine \cite{frantzi2000automatic} are examples of 
online systems for term extraction and rely on rule-based hybrid approaches. Finally, MultiTerm Extract\footnote{\url{https://www.trados.com/de/products/multiterm-desktop/}} and SketchEngine\footnote{\url{https://sketchengine.eu}} are available 
commercial systems.

\section{Conclusion}

In this paper, we proposed \textsc{Vocab-Expander}, an online tool that enables end-users to create and expand a vocabulary of their domain of interest. It uses state-of-the-art word embedding techniques based on web text as well as ConceptNet, a common-sense knowledge base, to suggest related terms for already given terms. The system can be used for a variety of purposes such as improving information retrieval, communication and collaboration, creating vocabularies for education, and fine-tuning language models in natural language processing.

For the future, we will 
allow for the integration of
domain-specific text corpora (e.g., provided by the domain experts) and provide a functionality to see to which degree the suggested terms occur in the text corpora.
Furthermore, we plan to evaluate the performance of \textsc{Vocab-Expander} by means of user studies in different domains and applications.

\section*{Acknowledgments}

This work was partially supported by the German Federal Ministry of Education and Research (BMBF) as part of the project IIDI (01IS21026D).

\bibliographystyle{acl_natbib}
\bibliography{bibliography}

\end{document}